\documentclass[10pt,twocolumn,letterpaper]{article}

\usepackage[pagenumbers]{iccv} 

\usepackage{booktabs}
\usepackage{colortbl}
\usepackage{caption}
\usepackage{amsmath}  
\usepackage{multirow}
\usepackage{graphicx}
\definecolor{bestval}{gray}{0.9}  

\definecolor{iccvblue}{rgb}{0.21,0.49,0.74}
\usepackage[pagebackref,breaklinks,colorlinks,allcolors=iccvblue]{hyperref}

\def\paperID{11} 
\def\confName{ICCV}
\def\confYear{2025}

\title{Data-Efficient Ensemble Weather Forecasting with Diffusion Models}

\author{%
Kevin Valencia \quad Ziyang Liu \quad Justin Cui\\
University of California, Los Angeles \\
{\tt\small \{kevinval04, jackieliu2025, justincui\}@ucla.edu}
}

\begin{document}
\maketitle

\begin{abstract}

Although numerical weather forecasting methods have dominated the field, recent advances in deep learning methods, such as diffusion models, have shown promise in ensemble weather forecasting. However, such models are typically autoregressive and are thus computationally expensive. This is a challenge in climate science, where data can be limited, costly, or difficult to work with. In this work, we explore the impact of curated data selection on these autoregressive diffusion models. We evaluate several data sampling strategies and show that a simple time stratified sampling approach achieves performance similar to or better than full-data training. Notably, it outperforms the full-data model on certain metrics and performs only slightly worse on others while using only 20\% of the training data. Our results demonstrate the feasibility of data-efficient diffusion training, especially for weather forecasting, and motivates future work on adaptive or model-aware sampling methods that go beyond random or purely temporal sampling.
\end{abstract}    
\section{Introduction}
\label{sec:intro}

In the field of climate science, weather forecasting and in particular ensemble forecasting has typically been conducted using numerical weather prediction (NWP) models. But in recent years, utilizing diffusion models has gained traction in ensemble weather forecasting. These models are known for their probabilistic generation capabilities and are trained in an autoregressive fashion. However, this fact makes them computationally demanding to train. At the same time, high-quality scientific data, especially within the climate and weather domains, can be limited or costly. Both of these constraints pose a significant barrier to the broader use of diffusion-based forecasting models when data is restricted.

While prior work in this field has largely focused on improving model architecture or forecasting accuracy, less attention has been paid to the role of data selection in training these models. Our aim is to explore the potential of curated data sampling to reduce training requirements. Our approach evaluates several simple sampling baselines under a fixed 20\% data budget. Our findings show that stratified time sampling, which ensures uniform representation across months, achieves performance close to full-data training across most metrics such as Continuous Ranked Probability Score (CRPS), Root Mean Squared Error (RMSE), and even surpasses it on Spread/Skill Ratio (SSR). 

\textbf{Our contribution:} We conduct the first focused evaluation of curated data sampling strategies for training autoregressive diffusion models in ensemble weather forecasting. We show that stratified time sampling, a simple temporal coverage heuristic, achieves performance comparable to full-data training while using only 20\% of the data and even outperforms it on SSR. We present this setup as a strong baseline for future work on adaptive, model-aware data selection strategies in  machine learning ensemble weather forecasting.

\section{Related works}
\label{sec:Related works}


\subsection{Diffusion Models}

Diffusion models are a class of generative models that learn to synthesize data by reversing a gradual noise corruption process \cite{ho2020denoising}. Originally developed for image generation, they have achieved state-of-the-art results in high-fidelity synthesis across domains such as text-to-image \cite{nichol2021glide}, image-to-video \cite{ho2022video}, and 3D generation. These models are typically trained to predict either noise or denoised samples at each step of a predefined diffusion process, and sample generation is carried out by iteratively denoising from a Gaussian prior. 

Recent advances have improved the training stability and sampling efficiency of diffusion models using techniques such as improved noise schedules \cite{karras2022elucidating}, score-based modeling \cite{song2020score}, and accelerated samplers. Due to their ability to model complex, multi-modal distributions and produce calibrated uncertainty estimates, diffusion models can have a wide range of scientific domain applications.

\subsection{Diffusion Models for Ensemble Forecasting}
NWP has long been the backbone of operational forecasting, but it remains computationally expensive and heavily reliant on physical parameterizations that are difficult to tune and generalize \cite{bauer2020ecmwf}. Recently, deep learning methods have emerged as fast and flexible alternatives, offering the potential to reduce inference time while maintaining accuracy such as Pangu-Weather \cite{Bi2023AccurateMG}. Among these, diffusion models like GenCast \cite{price2025probabilistic} or LADCast have shown particular promise for ensemble forecasting due to their ability to generate diverse, calibrated samples.
GenCast demonstrates improved probabilistic calibration and perceptual quality compared to deterministic baselines \cite{price2025probabilistic}. In addition, there has been work to further improve these models architecturally such as LaDCast \cite{zhuang2025ladcast}. Complementing this, recent work by \citet{andrae2025continuous} proposes a continuous-time diffusion model for ensemble weather forecasting that allows generation at arbitrary lead times. These methods suggest that diffusion models are well suited for data-driven ensemble forecasting.

\subsection{Data Selection and Coreset Methods}

Selecting informative subsets of data has been studied extensively in the context of coreset selection, curriculum learning, and training-data pruning. A recent survey by \citet{moser2025coreset} outlines a wide range of approaches, from geometry-based selection (e.g., clustering, herding) to training-dynamics-aware methods based on gradient similarity, influence functions, or loss trajectories. By reducing the number of training examples without significantly sacrificing performance, these methods can substantially improve training efficiency. Such efficiency reduces computational costs, improving the sustainaiblity of these models and reducing their impact on the environment \cite{wu2022sustainable}. While these techniques have shown promise in reducing training cost for image classification and language modeling, they remain largely unexplored in climate or spatiotemporal forecasting domains.

Recent methods like ICONS \cite{wu2024icons} and COINCIDE \cite{lee2024concept} use model-aware selection to retain performance with only a fraction of data in vision-language settings. However, such methods are rarely applied in scientific domains, where pretrained models may be unavailable and training dynamics are less stable.

\subsection{Dataset Distillation}
Dataset distillation aims to synthesize a small dataset that, when used to train a model, yields performance comparable to training on the full dataset. \citet{wang2018dataset} introduced this foundational idea, treating distilled images as hyperparameters optimized to match a target model's final performance. Subsequent work improved fidelity and generality of synthetic datasets using gradient matching \cite{zhao2020dataset}, label learning \cite{bohdal2020flexible}, differentiable augmentations \cite{zhao2021dataset} or via trajectory matching methods \cite{cazenavette2022dataset}.

In contrast, our work focuses on simple, static heuristics and evaluates their effectiveness in a data-scarce, domain-specific setting. Despite the simplicity of these strategies, we find they can approach or even exceed full-data performance on some forecasting metrics, motivating future work on model-aware sampling tailored to scientific domains.


\section{Methods}

\subsection{Task and Dataset}

We study the task of global ensemble weather forecasting using data from the ERA5 reanalysis dataset \cite{rasp2020weatherbench}. Specifically, we focus on predicting atmospheric and surface-level variables at future time steps based on current and past conditions. Forecasts are produced at 24-hour intervals for lead times ranging from 0 to 240 hours (0 to 10 days).

Our experiments use the downsampled version of ERA5 provided by the WeatherBench benchmark \cite{rasp2020weatherbench} which offers global coverage at $5.625^\circ$ spatial resolution and 1-hour temporal increments. The 5 main variables we focus on are: geopotential at 500 hPa ($z_{500}$), temperature at 850 hPa ($t_{850}$), 2-meter surface temperature ($t2m$), and the 10-meter zonal and meridional wind components ($u_{10}$ and $v_{10}$). These variables capture both mid-atmosphere dynamics and near-surface conditions, making them suitable for our weather forecasting task.
We preprocess the data exactly as \citet{andrae2025continuous} does. All variables are standardized by removing the training set mean and dividing by the standard deviation In addition to dynamic inputs, we include static fields such as the land-sea mask and orography, both rescaled to lie in the $[0, 1]$ range. The dataset is split by year: 1979–2015 for training, 2016–2017 for validation, and 2018 for testing. Forecasts are initialized from every hour of the year, excluding the first 24 hours and final 10 days to avoid overlap across splits.

\subsection{Baseline Methods} \label{sec:sampling}
Using autoregressive diffusion models for ensemble weather forecasting can become challenging and inefficient when it comes to collecting, storing, and using large-scale data. Motivated by this, we evaluate the impact of various data selection methods on the performance of such diffusion models. Specifically, we compare selection methods under a fixed training budget of 20\% of the full training set. Our aim is not to propose a new sampling method, but to benchmark simple baselines and assess whether targeted selection can approach or even exceed full-data performance.

\textbf{Full-Data.} This baseline uses the entire training period from 1979–2015 and serves as the upper bound in our comparisons.

\textbf{Random.} This baseline samples 20\% of the training data uniformly at random. While straightforward, it is sensitive to temporal autocorrelation in weather data: samples drawn from nearby time points may overrepresent certain weather phenomena while missing others entirely. As a result, random sampling can occasionally match stronger baselines but often shows high variance and inconsistent performance.

\textbf{K-Means.} To obtain a representative 20\% training subset, we apply dimensionality reduction followed by clustering in feature space. First, we flatten each training example and extract a feature matrix of shape $(N, D)$, where N is the length of the full data. We then apply Principal Component Analysis to reduce this to $M$ dimensions ($M \ll D$), capturing the dominant structure in the dataset. KMeans clustering is applied to the PCA-transformed data with $k = 0.2 \times N$ clusters. From each cluster, we select the data point closest to the centroid. This procedure ensures broad coverage across distinct data modes while avoiding redundancy. Unlike random sampling, this method provides a systematic and low-dimensional summary of the training distribution.

\textbf{Greedy Diverse Sampling.}
This method selects a subset of data points that are maximally diverse in spatial structure. We compute a distance metric (e.g., Euclidean distance between spatially averaged fields) and use a greedy algorithm to iteratively select the next example that is most dissimilar from those already chosen. The goal is to maximize the coverage of the training subset across distinct atmospheric configurations. Unlike K-means, this method does not impose global clusters, but instead builds a diverse set point by point. However, it can be sensitive to outliers and local variations.

\textbf{Stratified Time Sampling.} This method selects a fixed number of training samples uniformly from each calendar month across the full training period. This ensures that each subset includes a diverse range of seasonal conditions, such as winter storms, summer heatwaves, and transitional weather patterns. Unlike random sampling, which may overrepresent certain months or weather regimes due to temporal autocorrelation, stratified sampling provides balanced exposure to climatological variability.

While the method is simple and does not explicitly optimize for diversity or representativeness in feature space, it is guided by a strong domain prior: weather is highly seasonal, and adequate training across all months is crucial for generalization. In our experiments, stratified time sampling performs consistently well across metrics and variables, and even outperforms full-data training in forecast skill (SSR) on certain targets. These results suggest that temporally balanced sampling is a surprisingly effective strategy for data-efficient training in ensemble forecasting.

\textbf{Other Baselines.} We also evaluate spatial stratification, and herding-based subset matching. While these methods typically underperform compared to stratified time sampling on SSR, they often achieve RMSE and CRPS values within a small margin of the full-data baseline. Overall, the results suggest that many data-efficient models are competitive even when trained on small but representative subsets.

\section{Results}

\subsection{Model and Training Setup}
Our experiments use a fixed diffusion model architecture across all sampling strategies. The model is trained to forecast the weather state 24 hours into the future, given the current state. This setup mirrors the AR-24h configuration described in prior work \cite{price2025probabilistic,andrae2025continuous}, where a single-step 24-hour-ahead model is rolled out autoregressively during inference to produce multi-day forecasts. The model is conditioned on dynamic input fields and static geographic features (land-sea mask and orography), all of which are standardized or normalized as described in the previous section. Across all experiments, we keep the model architecture, training objective, and hyperparameters fixed. The only variable is the subset of training data used, which is determined by the sampling strategy. This allows us to isolate the effect of data selection on forecasting performance. 
We evaluate all sampling methods on the 2018 test year using the metrics described in the next section. Forecasts are generated autoregressively up to 10 days in 24-hour increments, and metrics are computed at 5-day and 10-day lead times.

\subsection{Evaluation Metrics}
We evaluate forecasting performance using three standard metrics from the weather and climate modeling literature: RMSE, CRPS, and SSR. These metrics are computed over the forecast lead times of interest, at 5 and 10 days.

\textbf{RMSE.} Measures the average difference between the ensemble mean forecast and the ground truth. It captures overall predictive accuracy and is sensitive to large errors.

\textbf{CRPS.} A probabilistic scoring rule that generalizes RMSE to full distributions. It compares the predicted ensemble distribution to the true outcome, rewarding both accuracy and sharpness \cite{hersbach2000decomposition}.

\textbf{SSR.} The Spread/Skill Ratio is a calibration metric that compares ensemble spread to forecast error. Ideally, SSR should be close to 1, indicating that the model's uncertainty estimates are well-calibrated \cite{fortin2014should}.

\subsection{Main Results}

Our key finding is that stratified time sampling achieves comparable or better performance than full-data training on SSR, despite using only 20\% of the training data. For all metrics on both 5-day and 10-day lead times this stratified time outperforms or matches the full-data model in SSR,  while maintaining close RMSE and CRPS.

\begin{table}[ht]
\centering
\resizebox{\columnwidth}{!}{
\begin{tabular}{lcccccc}
\toprule
\textbf{Sampling Method} & \multicolumn{2}{c}{\textbf{CRPS}} & \multicolumn{2}{c}{\textbf{RMSE}} & \multicolumn{2}{c}{\textbf{SSR}} \\
& 5 days & 10 days & 5 days & 10 days & 5 days & 10 days \\
\midrule
Full Data     & \textbf{242.66} & \textbf{335.2}  & \textbf{544.19} & \textbf{750.52} & 0.84 & 0.94 \\
Greedy Diverse         & 265.07 & 363.61  & 575.10 & 782.72 & 0.87 & 0.95 \\
Herding                & 289.10 & 379.70  & 609.33 & 807.61 & 0.70 & 0.88 \\
KMeans                 & 264.35 & 366.67  & 570.01 & 784.81 & 0.85 & 0.93 \\
Random                 & 267.02 & 368.76  & 571.24 & 782.21 & 0.85 & 0.95 \\
Spatial                & 264.90 & 361.38  & 579.11 & 788.23 & 0.86 & 0.94 \\
Stratified Time            & \underline{257.58} & \underline{360.2}  & \underline{560.87} & \underline{779.36} & \textbf{0.89} & \textbf{0.97} \\
\bottomrule
\end{tabular}
}
\caption{Forecast performance for $z_{500}$ at 5-day/10-day lead times. Lower is better for RMSE and CRPS; closer to 1 is better for SSR.}
\end{table}

\begin{table}[ht]
\centering
\resizebox{\columnwidth}{!}{
\begin{tabular}{lcccccc}
\toprule
\textbf{Sampling Method} & \multicolumn{2}{c}{\textbf{CRPS}} & \multicolumn{2}{c}{\textbf{RMSE}} & \multicolumn{2}{c}{\textbf{SSR}} \\
& 5 days & 10 days & 5 days & 10 days & 5 days & 10 days \\
\midrule
Full Data & \textbf{1.24} & \textbf{1.6} & \textbf{2.55} & \textbf{3.25} & 0.89 & 0.96 \\
Greedy Diverse     & 1.36 & 1.73 & 2.69 & 3.38 & 0.92 & 0.96 \\
Herding            & 1.44 & 1.79 & 2.82 & 3.47 & 0.79 & 0.90 \\
KMeans             & 1.35 & 1.74 & 2.67 & 3.39 & 0.90 & 0.95 \\
Random             & 1.36 & 1.74 & 2.68 & 3.38 & 0.90 & 0.95 \\
Spatial            & 1.38 & 1.76 & 2.72 & 3.42 & 0.90 & 0.94 \\
Stratified Time        & \underline{1.33} & \underline{1.72} & \underline{2.64} & \underline{3.37} & \textbf{0.93} & \textbf{0.98} \\
\bottomrule
\end{tabular}
}
\caption{Forecast performance for $t_{850}$ at 5-day/10-day lead times.}
\end{table}


\begin{figure}[ht]
    \centering
    \includegraphics[width=0.73\columnwidth,
                     trim=10pt 8pt 12pt 10pt, clip]{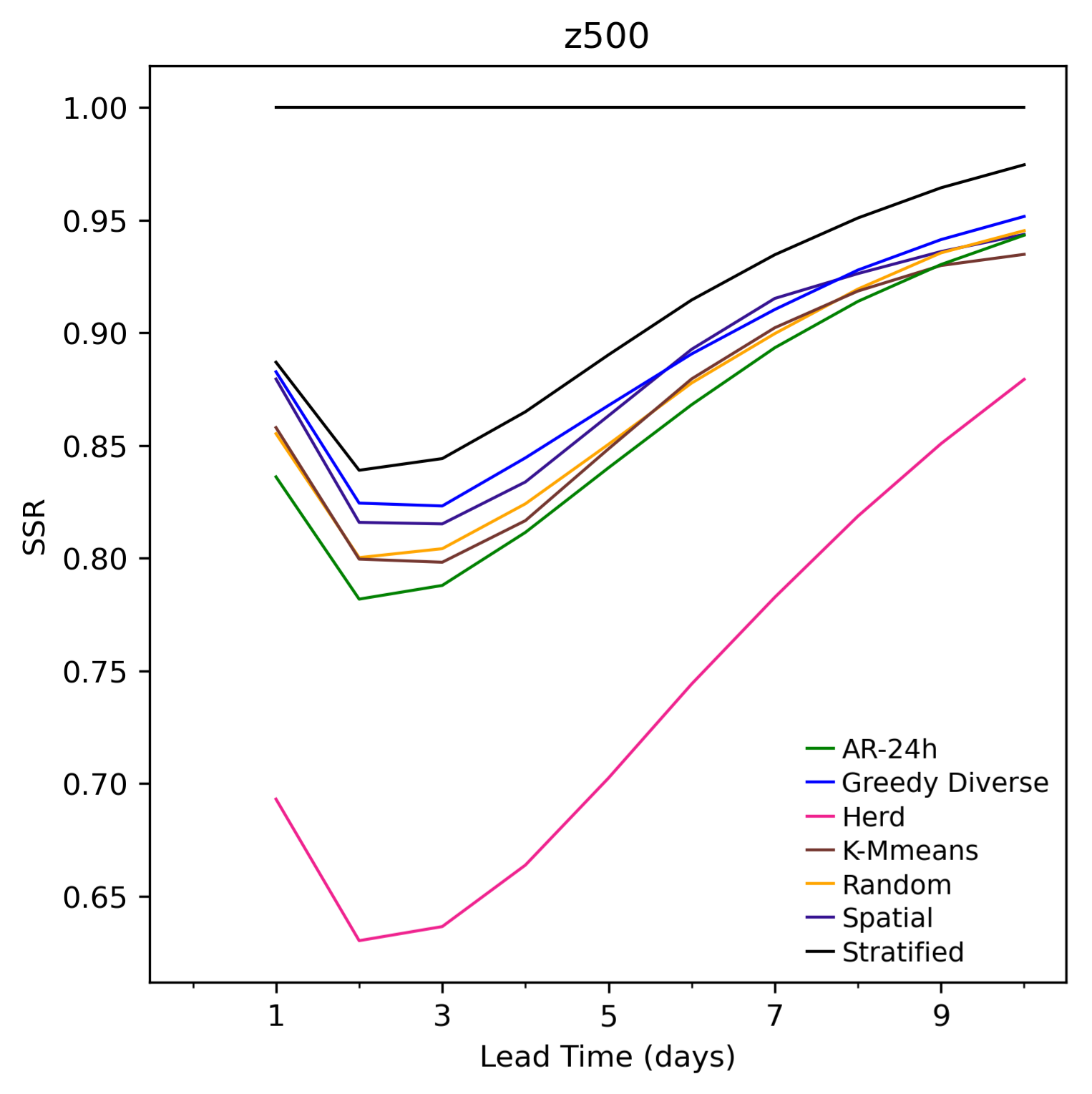}
    \caption{SSR across sampling methods for $z_{500}$. Stratified time performs best.}
    \label{fig:ssr_plot}
\end{figure}

\subsection{Sampling Discussion}

Stratified time sampling performs best among the 20\% data baselines, likely due to its coverage across all months in the training period. By ensuring that each season is represented across the 12 months, the model is exposed to a more balanced variety of atmospheric phenomena (e.g., winter storms, summer heatwaves). In contrast, other methods such as random or spatial sampling may inadvertently over-sample from particular times of year, leading to seasonal biases and poorer generalization.

Across all baselines, we observe that RMSE and CRPS remain relatively stable since most models perform within a relatively narrow margin of the full-data baseline for each variable. In contrast, SSR shows larger variation, with some methods producing significantly underdispersed ensembles. This suggests that sampling diversity plays a more critical role in calibrating uncertainty estimates than in improving point prediction accuracy. As SSR depends on the alignment between forecast error and spread, models trained on more representative data subsets are better able to learn uncertainty structure.

\subsection{Findings on Other Variables}

We include full results for $t2m$, $u_{10}$, $v_{10}$, and $ws_{10}$ in Appendix~\ref{app:Full_Results}. These follow similar trends: stratified time performs consistently well, and most 20\% subsets yield reasonable accuracy despite the data reduction. However, surface variables tend to show slightly larger performance gaps, likely due to their greater local variability.

\section{Discussion and Future Work}
Our experiments show that simple data sampling strategies like stratified time sampling can enable strong performance in diffusion-based ensemble weather forecasting, even when using only 20\% of the training data. Stratified time sampling consistently achieves comparable CRPS and RMSE to full-data training and even outperforms it on SSR in several cases. This result is especially notable given the simplicity of the method, which selects a fixed number of samples uniformly from each calendar month. At the same time, our findings highlight important limitations and opportunities. All sampling baselines we evaluate are manually designed, static, and not data-driven. Additionally, our experiments are limited to a single test year (2018), which may not capture longer-term variability. Still, the fact that such simple, non-adaptive sampling methods can match or even exceed full-data performance in some cases with only 20\% of the original training data, strongly motivates further exploration in this direction of curated data subsets.


\section{Conclusion}
In this work, we present a first study on data-efficient training for autoregressive diffusion models in ensemble weather forecasting. By evaluating several sampling strategies under a fixed 20\% data budget, we show that simple heuristics, in particular stratified time sampling, can achieve performance comparable to full-data training, and in most cases even improve forecast skill. These findings suggest that thoughtful data selection can substantially reduce training cost without sacrificing quality. Our results provide a foundation for future work on adaptive, model-aware sampling strategies for scientific forecasting domains.
\clearpage

{
    \small
    \bibliographystyle{ieeenat_fullname}
    \bibliography{main}

\begin{thebibliography}{22}
\providecommand{\natexlab}[1]{#1}
\providecommand{\url}[1]{\texttt{#1}}
\expandafter\ifx\csname urlstyle\endcsname\relax
  \providecommand{\doi}[1]{doi: #1}\else
  \providecommand{\doi}{doi: \begingroup \urlstyle{rm}\Url}\fi

\bibitem[Andrae et~al.(2025)Andrae, Landelius, Oskarsson, and
  Lindsten]{andrae2025continuous}
Martin Andrae, Tomas Landelius, Joel Oskarsson, and Fredrik Lindsten.
\newblock Continuous ensemble weather forecasting with diffusion models.
\newblock In \emph{The Thirteenth International Conference on Learning
  Representations}, 2025.

\bibitem[Bauer et~al.(2020)Bauer, Quintino, Wedi, Bonanni, Chrust, Deconinck,
  Diamantakis, D{\"u}ben, English, Flemming, et~al.]{bauer2020ecmwf}
Peter Bauer, Tiago Quintino, Nils Wedi, Antonio Bonanni, Marcin Chrust, Willem
  Deconinck, Michail Diamantakis, Peter D{\"u}ben, Stephen English, Johannes
  Flemming, et~al.
\newblock \emph{The ECMWF scalability programme: Progress and plans}.
\newblock European Centre for Medium Range Weather Forecasts, 2020.

\bibitem[Bi et~al.(2023)Bi, Xie, Zhang, Chen, Gu, and Tian]{Bi2023AccurateMG}
Kaifeng Bi, Lingxi Xie, Hengheng Zhang, Xin Chen, Xiaotao Gu, and Qi Tian.
\newblock Accurate medium-range global weather forecasting with 3d neural
  networks.
\newblock \emph{Nature}, 619:\penalty0 533 -- 538, 2023.

\bibitem[Bohdal et~al.(2020)Bohdal, Yang, and Hospedales]{bohdal2020flexible}
Ondrej Bohdal, Yongxin Yang, and Timothy Hospedales.
\newblock Flexible dataset distillation: Learn labels instead of images.
\newblock \emph{arXiv preprint arXiv:2006.08572}, 2020.

\bibitem[Cazenavette et~al.(2022)Cazenavette, Wang, Torralba, Efros, and
  Zhu]{cazenavette2022dataset}
George Cazenavette, Tongzhou Wang, Antonio Torralba, Alexei~A Efros, and
  Jun-Yan Zhu.
\newblock Dataset distillation by matching training trajectories.
\newblock In \emph{Proceedings of the IEEE/CVF Conference on Computer Vision
  and Pattern Recognition}, pages 4750--4759, 2022.

\bibitem[Fortin et~al.(2014)Fortin, Abaza, Anctil, and
  Turcotte]{fortin2014should}
Vincent Fortin, Mabrouk Abaza, Francois Anctil, and Raphael Turcotte.
\newblock Why should ensemble spread match the rmse of the ensemble mean?
\newblock \emph{Journal of Hydrometeorology}, 15\penalty0 (4):\penalty0
  1708--1713, 2014.

\bibitem[Hersbach(2000)]{hersbach2000decomposition}
Hans Hersbach.
\newblock Decomposition of the continuous ranked probability score for ensemble
  prediction systems.
\newblock \emph{Weather and Forecasting}, 15\penalty0 (5):\penalty0 559--570,
  2000.

\bibitem[Ho et~al.(2020)Ho, Jain, and Abbeel]{ho2020denoising}
Jonathan Ho, Ajay Jain, and Pieter Abbeel.
\newblock Denoising diffusion probabilistic models.
\newblock \emph{Advances in neural information processing systems},
  33:\penalty0 6840--6851, 2020.

\bibitem[Ho et~al.(2022)Ho, Salimans, Gritsenko, Chan, Norouzi, and
  Fleet]{ho2022video}
Jonathan Ho, Tim Salimans, Alexey Gritsenko, William Chan, Mohammad Norouzi,
  and David~J Fleet.
\newblock Video diffusion models.
\newblock \emph{Advances in Neural Information Processing Systems},
  35:\penalty0 8633--8646, 2022.

\bibitem[Karras et~al.(2022)Karras, Aittala, Aila, and
  Laine]{karras2022elucidating}
Tero Karras, Miika Aittala, Timo Aila, and Samuli Laine.
\newblock Elucidating the design space of diffusion-based generative models.
\newblock \emph{Advances in neural information processing systems},
  35:\penalty0 26565--26577, 2022.

\bibitem[Lee et~al.(2024)Lee, Li, and Hwang]{lee2024concept}
Jaewoo Lee, Boyang Li, and Sung~Ju Hwang.
\newblock Concept-skill transferability-based data selection for large
  vision-language models.
\newblock \emph{arXiv preprint arXiv:2406.10995}, 2024.

\bibitem[Moser et~al.(2025)Moser, Shanbhag, Frolov, Raue, Folz, and
  Dengel]{moser2025coreset}
Brian~B Moser, Arundhati~S Shanbhag, Stanislav Frolov, Federico Raue, Joachim
  Folz, and Andreas Dengel.
\newblock A coreset selection of coreset selection literature: Introduction and
  recent advances.
\newblock \emph{arXiv preprint arXiv:2505.17799}, 2025.

\bibitem[Nichol et~al.(2021)Nichol, Dhariwal, Ramesh, Shyam, Mishkin, McGrew,
  Sutskever, and Chen]{nichol2021glide}
Alex Nichol, Prafulla Dhariwal, Aditya Ramesh, Pranav Shyam, Pamela Mishkin,
  Bob McGrew, Ilya Sutskever, and Mark Chen.
\newblock Glide: Towards photorealistic image generation and editing with
  text-guided diffusion models.
\newblock \emph{arXiv preprint arXiv:2112.10741}, 2021.

\bibitem[Price et~al.(2025)Price, Sanchez-Gonzalez, Alet, Andersson, El-Kadi,
  Masters, Ewalds, Stott, Mohamed, Battaglia, et~al.]{price2025probabilistic}
Ilan Price, Alvaro Sanchez-Gonzalez, Ferran Alet, Tom~R Andersson, Andrew
  El-Kadi, Dominic Masters, Timo Ewalds, Jacklynn Stott, Shakir Mohamed, Peter
  Battaglia, et~al.
\newblock Probabilistic weather forecasting with machine learning.
\newblock \emph{Nature}, 637\penalty0 (8044):\penalty0 84--90, 2025.

\bibitem[Rasp et~al.(2020)Rasp, Dueben, Scher, Weyn, Mouatadid, and
  Thuerey]{rasp2020weatherbench}
Stephan Rasp, Peter~D Dueben, Sebastian Scher, Jonathan~A Weyn, Soukayna
  Mouatadid, and Nils Thuerey.
\newblock Weatherbench: a benchmark data set for data-driven weather
  forecasting.
\newblock \emph{Journal of Advances in Modeling Earth Systems}, 12\penalty0
  (11):\penalty0 e2020MS002203, 2020.

\bibitem[Song et~al.(2020)Song, Sohl-Dickstein, Kingma, Kumar, Ermon, and
  Poole]{song2020score}
Yang Song, Jascha Sohl-Dickstein, Diederik~P Kingma, Abhishek Kumar, Stefano
  Ermon, and Ben Poole.
\newblock Score-based generative modeling through stochastic differential
  equations.
\newblock \emph{arXiv preprint arXiv:2011.13456}, 2020.

\bibitem[Wang et~al.(2018)Wang, Zhu, Torralba, and Efros]{wang2018dataset}
Tongzhou Wang, Jun-Yan Zhu, Antonio Torralba, and Alexei~A Efros.
\newblock Dataset distillation.
\newblock \emph{arXiv preprint arXiv:1811.10959}, 2018.

\bibitem[Wu et~al.(2022)Wu, Raghavendra, Gupta, Acun, Ardalani, Maeng, Chang,
  Aga, Huang, Bai, et~al.]{wu2022sustainable}
Carole-Jean Wu, Ramya Raghavendra, Udit Gupta, Bilge Acun, Newsha Ardalani,
  Kiwan Maeng, Gloria Chang, Fiona Aga, Jinshi Huang, Charles Bai, et~al.
\newblock Sustainable ai: Environmental implications, challenges and
  opportunities.
\newblock \emph{Proceedings of Machine Learning and Systems}, 4:\penalty0
  795--813, 2022.

\bibitem[Wu et~al.(2024)Wu, Xia, Shao, Deng, Koh, and Russakovsky]{wu2024icons}
Xindi Wu, Mengzhou Xia, Rulin Shao, Zhiwei Deng, Pang~Wei Koh, and Olga
  Russakovsky.
\newblock Icons: Influence consensus for vision-language data selection.
\newblock \emph{arXiv preprint arXiv:2501.00654}, 2024.

\bibitem[Zhao and Bilen(2021)]{zhao2021dataset}
Bo Zhao and Hakan Bilen.
\newblock Dataset condensation with differentiable siamese augmentation.
\newblock In \emph{International Conference on Machine Learning}, pages
  12674--12685. PMLR, 2021.

\bibitem[Zhao et~al.(2020)Zhao, Mopuri, and Bilen]{zhao2020dataset}
Bo Zhao, Konda~Reddy Mopuri, and Hakan Bilen.
\newblock Dataset condensation with gradient matching.
\newblock \emph{arXiv preprint arXiv:2006.05929}, 2020.

\bibitem[Zhuang and Duraisamy(2025)]{zhuang2025ladcast}
Yilin Zhuang and Karthik Duraisamy.
\newblock Ladcast: A latent diffusion model for medium-range ensemble weather
  forecasting.
\newblock \emph{arXiv preprint arXiv:2506.09193}, 2025.

\end{thebibliography}
}

\clearpage
\setcounter{page}{1}
\maketitlesupplementary

\appendix
\section{Appendix}
\label{Appendix}

\subsection{Full Results}
\label{app:Full_Results}
Here we show the tables and results for our sampling methods on $t2m$, $u_{10}$, $v_{10}$, and $ws_{10}$. The performance of the sampling methods largely remain the same across all variables including the two from the main paper.

\begin{table}[ht]
\centering
\resizebox{\columnwidth}{!}{
\begin{tabular}{lcccccc}
\toprule
\textbf{Sampling Method} & \multicolumn{2}{c}{\textbf{CRPS}} & \multicolumn{2}{c}{\textbf{RMSE}} & \multicolumn{2}{c}{\textbf{SSR}} \\
& 5 days & 10 days & 5 days & 10 days & 5 days & 10 days \\
\midrule
Full Data & \textbf{0.87} & \textbf{1.09} & \textbf{1.98} & \textbf{2.48} & 0.90 & 0.95 \\
Greedy Diverse     & 0.97 & 1.19 & 2.12 & 2.60 & 0.93 & 0.97 \\
Herding            & 1.01 & 1.24 & 2.20 & 2.68 & 0.82 & 0.91 \\
KMeans             & 0.97 & 1.22 & 2.10 & 2.62 & 0.92 & 0.97 \\
Random             & 0.96 & 1.20 & 2.09 & 2.60 & 0.91 & 0.95 \\
Spatial            & 1.01 & 1.27 & 2.17 & 2.69 & 0.90 & 0.91 \\
Stratified Time    & \underline{0.94} & \underline{1.17} & \underline{2.06} & \underline{2.57} & \textbf{0.93} & \textbf{0.98} \\
\bottomrule
\end{tabular}
}

\caption{Forecast performance for $t2m$ at 5-day and 10-day lead times.}
\end{table}

\begin{table}[ht]
\centering
\resizebox{\columnwidth}{!}{
\begin{tabular}{lcccccc}
\toprule
\textbf{Sampling Method} & \multicolumn{2}{c}{\textbf{CRPS}} & \multicolumn{2}{c}{\textbf{RMSE}} & \multicolumn{2}{c}{\textbf{SSR}} \\
& 5 days & 10 days & 5 days & 10 days & 5 days & 10 days \\
\midrule
AR-24h (full data)    & \textbf{1.62} & \textbf{1.90} & \textbf{3.32} & \textbf{3.85} & 0.92 & 0.97 \\
Greedy Diverse        & 1.78 & 2.07 & 3.48 & 4.01 & 0.93 & 0.97 \\
Herding               & 1.86 & 2.11 & 3.61 & 4.08 & 0.82 & 0.93 \\
KMeans                & 1.76 & 2.06 & 3.47 & 4.00 & 0.93 & 0.97 \\
Random                & 1.76 & 2.07 & 3.46 & 4.01 & 0.93 & 0.97 \\
Spatial               & 1.78 & 2.08 & 3.49 & 4.02 & 0.93 & 0.97 \\
Stratified Time       & \underline{1.73} & \underline{2.05} & \underline{3.42} & \underline{3.99} & \textbf{0.94} & \textbf{0.98} \\
\bottomrule
\end{tabular}
}
\caption{Forecast performance for $u_{10}$ at 5-day and 10-day lead times.}
\end{table}

\begin{table}[ht]
\centering
\resizebox{\columnwidth}{!}{
\begin{tabular}{lcccccc}
\toprule
\textbf{Sampling Method} & \multicolumn{2}{c}{\textbf{CRPS}} & \multicolumn{2}{c}{\textbf{RMSE}} & \multicolumn{2}{c}{\textbf{SSR}} \\
& 5 days & 10 days & 5 days & 10 days & 5 days & 10 days \\
\midrule
AR-24h (full data)    & \textbf{1.68} & \textbf{1.96} & \textbf{3.42} & \textbf{3.96} & 0.93 & \textbf{0.99} \\
Greedy Diverse        & 1.84 & 2.13 & 3.58 & 4.12 & 0.94 & 0.98 \\
Herding               & 1.91 & 2.16 & 3.70 & 4.17 & 0.83 & 0.94 \\
KMeans                & 1.82 & 2.12 & 3.56 & 4.11 & 0.93 & 0.98 \\
Random                & 1.82 & 2.12 & 3.56 & 4.12 & 0.94 & 0.98 \\
Spatial               & 1.84 & 2.13 & 3.60 & 4.13 & 0.94 & 0.98 \\
Stratified Time       & \underline{1.79} & \underline{2.11} & \underline{3.51} & \underline{4.10} & \textbf{0.95} & \textbf{0.99} \\
\bottomrule
\end{tabular}
}
\caption{Forecast performance for $v_{10}$ at 5-day and 10-day lead times.}
\end{table}

For some variables, some sampling methods seemed to perform better than on others. However, stratified time consistently provided strong results across most variables, particularly in SSR. This aligns with our intuition that temporal representativeness is a strong prior in weather-related forecasting tasks. We also provide the SSR plot across the sampling methods for $t_{850}$ which has similar results to $z_{500}$ from \ref{fig:ssr_plot}.

\begin{figure}[ht]
    \centering
    \includegraphics[width=0.85\columnwidth]{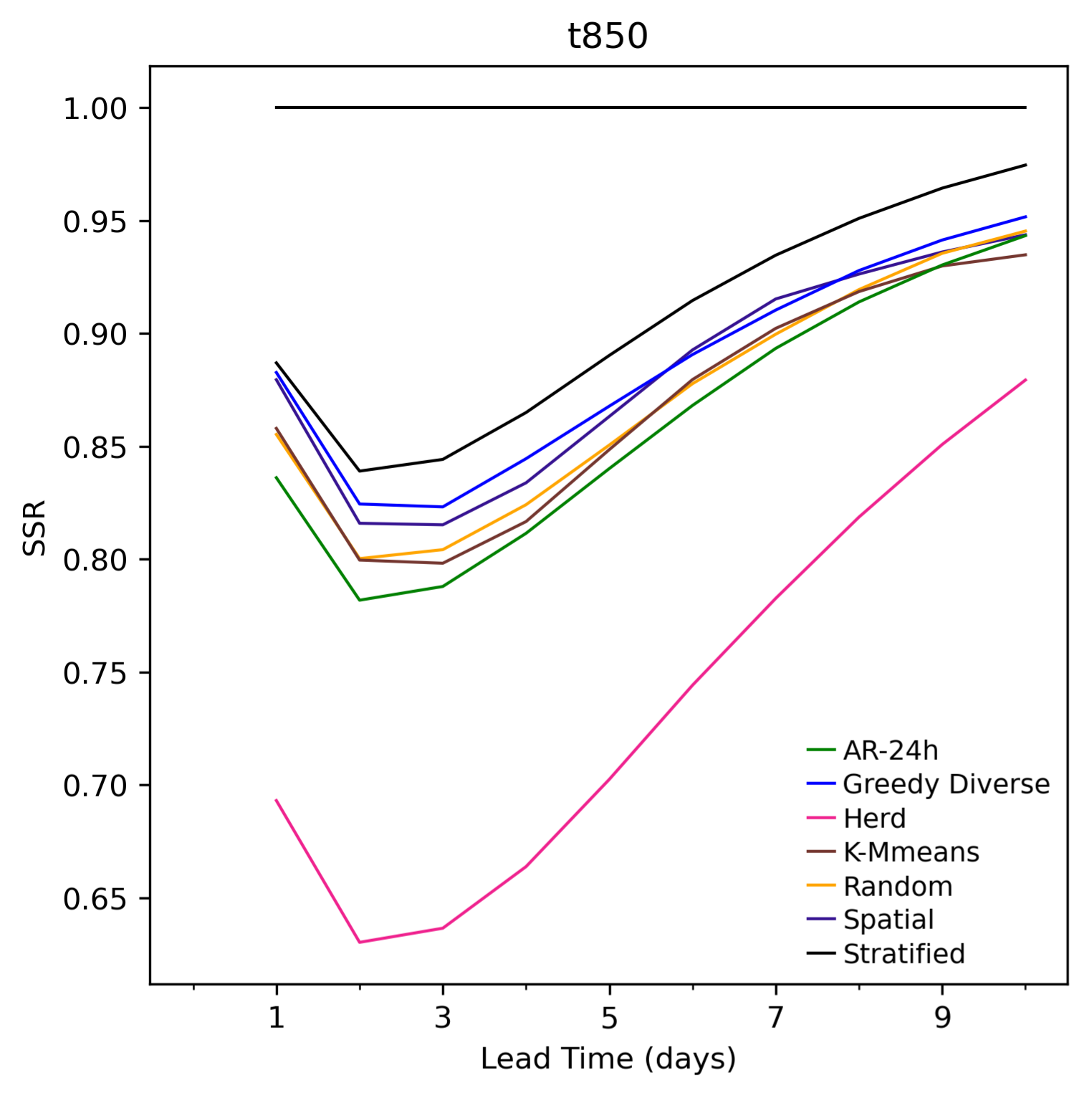}
    \caption{SSR across different sampling methods for $t_{850}$. Stratified time performs best.}
    \label{fig:ssr_plot2}
\end{figure}

\subsection{Extended Sampling Variants}
\label{app:extended_sampling}

In addition to standard baselines, we experimented with several hybrid sampling methods motivated by the success of stratified time sampling. These include:

\begin{itemize}
    \item \textbf{Stratified KMeans:} Combines stratified time sampling with spatial clustering. We first divide the dataset uniformly across months, then within each month, apply k-means clustering on spatial statistics to ensure diversity.
    
    \item \textbf{Stratified Entropy:} Selects samples with high predictive uncertainty within each month. This encourages selection of "harder" or more informative samples while maintaining temporal uniformity.

    \item \textbf{Stratified Spatial Diversity:} Within each month, selects samples to maximize pairwise spatial dissimilarity, measured by cosine distance between spatial mean vectors.

    \item \textbf{Stratified KMeans++:} Uses k-means++ initialization instead of standard k-means within monthly bins to improve cluster representativeness.

\end{itemize}

Each of these variants was motivated by the desire to combine the robustness and simplicity of stratified time sampling with additional structure. Performance results however were lacking and thus were not included in the main sections. These experiments nevertheless provided valuable insights into the limitations of combining multiple heuristics, suggesting that more principled or learned approaches to hybrid sampling may be a promising direction for future work.

\end{document}